\documentclass[11pt]{article}
\usepackage[margin=1in]{geometry}
\usepackage[utf8]{inputenc}
\usepackage{microtype}
\usepackage{amsmath,amssymb,mathtools}
\usepackage{graphicx}
\usepackage{booktabs}
\usepackage{xcolor}
\usepackage{fancyhdr}
\usepackage{hyperref}
\hypersetup{
  colorlinks=true,
  linkcolor=blue!70!black,
  citecolor=green!50!black,
  urlcolor=blue!80!black,
  bookmarksnumbered=true,
  pdfauthor={Ioannis Tsiokos},
  pdftitle={To Throw a Stone with Six Birds: On Agents and Agenthood}
}
\usepackage{enumitem}
\usepackage{natbib}

\newcommand{\SBT}{Six Birds Theory}
\newcommand{\Emp}{\mathrm{Emp}}
\newcommand{\K}{\mathcal{K}}
\newcommand{\Alpha}{\mathcal{A}}
\DeclareUnicodeCharacter{03B1}{\ensuremath{\alpha}}
\DeclareUnicodeCharacter{2200}{\ensuremath{\forall}}
\DeclareUnicodeCharacter{2203}{\ensuremath{\exists}}
\DeclareUnicodeCharacter{2192}{\ensuremath{\to}}
\DeclareUnicodeCharacter{2286}{\ensuremath{\subseteq}}
\DeclareUnicodeCharacter{2227}{\ensuremath{\wedge}}

\title{To Throw a Stone with Six Birds:\\ On Agents and Agenthood}
\author{Ioannis Tsiokos\\\texttt{ioannis@automorph.io}}
\date{31 January 2026}

\fancypagestyle{firstpage}{%
  \fancyhf{}
  \fancyfoot[C]{\scriptsize DOI: \href{https://doi.org/10.5281/zenodo.18439737}{10.5281/zenodo.18439737} \quad \textcopyright\ 2026 Automorph Inc.}

}

\begin{document}
\maketitle
\thispagestyle{firstpage}

\begin{abstract}
In \SBT\ (\citep{six_birds_theory}), macroscopic objects are induced by layers/closures, yet empirical discussions of agency often conflate persistence (being an object) with control (making a difference), making agency claims hard to verify and easy to fake. We propose a type-correct \SBT\ account: a theory is an induced layer $T=(\Pi,L,\mathcal F,B)$; an agent is a \emph{theory object}---a maintained package inside $T$ with a ledger-gated interface. Operationally, we instantiate this contract as exact finite controlled kernels with (i) ledger-gated feasibility, (ii) robust viability kernels under successor-support semantics, (iii) feasible empowerment (channel capacity) as a difference-making proxy, and (iv) an empirical packaging endomap whose idempotence defect measures objecthood for coarse lenses. Across matched-control ablations in a minimal ring-world, we obtain four checkable separations: (1) in calibrated nulls, single-action regimes have 0-bit empowerment and an exogenous-schedule trap yields 1 bit only under the wrong model (0 under the correct one); (2) enabling repair collapses idempotence defect at $\tau=2$ from 1 to 0; (3) enabling protocol holonomy leaves $H=1$ unchanged but increases empowerment for $H\ge2$; and (4) operator rewriting (skill $\theta$) monotonically increases median empowerment (0.73$\to$1.34 bits). These results separate agenthood from agency with audited, hash-traceable artifacts. We are explicit that stronger claims about goals, consciousness, or real organisms are not established here; our conclusions rely on finite witnesses and operational proxies under explicit controls.
\end{abstract}


\section{Introduction}
\label{sec:introduction}

\subsection{To throw a stone}
To \emph{throw a stone} is to make a reliable difference in the outside world: after the throw, the world is measurably different than it would have been otherwise. That sounds trivial until one asks what must already be true for the sentence to be meaningful. Where is the ``throw'' located in a messy substrate of micro-interactions? Which degrees of freedom count as the thrower, which as the stone, and which as the outside world? When does a subsystem have \emph{choices} rather than merely being pushed around?

This paper treats agency as an \emph{emergent, induced} notion. The stone-throwing motif is a reminder that agency is not primarily about internal stories (goals, intentions, preferences), but about \emph{counterfactual difference-making under constraints}: there exist at least two feasible interface policies that lead to measurably different outside futures at the relevant horizon.

\subsection{From objects to agents in \SBT}
In \SBT\ (\citep{six_birds_theory}), macroscopic objects are not assumed as primitives. Instead, \SBT\ proposes six emergence primitives (``six birds'') that enable stable macroscopic descriptions: writing/closure, constraints, protocol holonomy, quantization/identity staging, viability closure, and resource transduction. On this view, an ``object'' is a stabilized package: a closure that persists long enough that it can be treated as a thing with boundary conditions.

A natural next step is agency. If objects are induced, then agents should be induced too. The Life paper (\citep{six_birds_life}) already pushes in this direction: life is treated as a maintenance-and-closure phenomenon rather than a list of biological exceptions. Here we ask the parallel question for agency: \emph{what must be enabled so that something like an action exists, and once enabled, what does it mean for that action to be causal?}

\subsection{Agenthood versus agency}
We separate two notions that are often conflated.

\paragraph{Agenthood (enablement across layers).}
Before there is an agent, there are only micro-interactions. An agent is a \emph{packaged closure} that becomes stable enough to support an internal/external distinction, a ledger, and a controllable interface. This is an enablement claim: it is about the existence and persistence of a layer in which ``actions'' are well-defined variables. In \SBT\ terms, agenthood is primarily built from packaging/viability (P$_5$), accounting (P$_6$), and feasibility constraints (P$_2$), typically with staging/identity (P$_4$) to set a meaningful horizon.

\paragraph{Agency (causation inside a layer).}
Once an agent layer exists, we can ask causal questions: do different feasible interface choices lead to different outside futures? This is a within-layer notion: interventions on an action variable should change the distribution of future outside macrostates. In our finite setting, we operationalize this ``difference-making'' using channel capacity (empowerment) restricted to feasible action sequences.

This separation matters because many apparent agency signals are artifacts of modeling. A single-action system can look active but has no control capacity; an exogenous schedule can be mistaken for choice if incorrectly treated as an action. We include explicit null regimes to guard against such false positives.

\subsection{Thesis: an agent is a \emph{theory object}}
Our thesis is: an agent is a \emph{theory object}. In \SBT, a \emph{theory} is a \emph{layer/closure}: an induced macro-physics (often written $T=(\Pi,L,\mathcal F,B)$) that makes certain macro-variables stable and well-typed for causal talk. A \emph{theory object} is not the layer itself, but an object \emph{in} that layer: a package that persists under the layer's closure and accounting rules, and whose boundary degrees of freedom can be intervened upon.

Concretely, in this paper the finite controlled kernel together with its lenses, feasibility gate, and ledger instantiates such a theory at a chosen scale. The \emph{agent} is the stabilized object this theory induces: a viable packaged subsystem (captured by a nonempty viability kernel and low packaging defect) equipped with an interface alphabet that supports a nontrivial feasible action channel to outside futures.

In \SBT\ terms, agenthood is the enablement of a causally thick layer; the agent is the corresponding theory object (a maintained invariant package) inside that layer. Agency is then the within-layer causal content: interventions on the interface change the distribution of outside macrostates at the relevant horizon.

\subsection{Operational plan and evidence}
We make the thesis testable by committing to a small, exact finite-state substrate and three complementary metrics.

\begin{enumerate}[leftmargin=*, itemsep=0.25em]
  \item \textbf{Viability as a greatest fixed point.} We compute a robust viability kernel $\K$ (greatest controlled-invariant safe set) under support semantics: a state remains viable if there exists a feasible action whose successor support stays inside $\K$.
  \item \textbf{Difference-making as feasible empowerment.} For a horizon $H$, we build the induced channel from action sequences to an outside output variable and compute its capacity, restricted by ledger-gated feasibility (budgets and costs).
  \item \textbf{Packaging stability as an empirical endomap.} For a macro lens that hides micro-degrees of freedom, we empirically define an endomap $E$ on macro labels by rolling forward and taking the mode. Its \emph{idempotence defect} quantifies whether the macro labels behave like stable objects under that lens.
\end{enumerate}

We instantiate these in a minimal ring-world environment with toggles corresponding to the \SBT\ primitives: protocol holonomy (P$_3$), repair/maintenance accounting (P$_6$), feasibility gating (P$_2$), optional identity sector (P$_4$), and optional operator rewriting (learning, P$_1$). The exhibits then show: repair collapses packaging defect; protocol produces horizon-dependent empowerment gains; learning increases empowerment with skill; and null regimes demonstrate how agency can be faked by mis-modeling.

\subsection{Guide to the paper}
Section~\ref{sec:dictionary} gives a compact dictionary from \SBT\ primitives to agency roles and to the metrics we use. Section~\ref{sec:engine} defines the finite computational substrate and the operational notions (viability, feasible empowerment, packaging endomap). Sections~\ref{sec:ex_packaging}--\ref{sec:ex_learning} present the quantitative exhibits (packaging, nulls, holonomy, ablations, sweep, learning). Section~\ref{sec:repro} documents reproducibility: hashed configs, audited artifacts, and a small Lean lemma anchoring viability as a greatest fixed point. We close in Section~\ref{sec:discussion} with implications, limitations, and how ``an agent is a \emph{theory object}'' scales beyond this toy substrate.

\paragraph{Code availability.}
\begin{itemize}[noitemsep,topsep=2pt]
\item Repository: \url{https://github.com/ioannist/six-birds-agent}
\end{itemize}

\subsection*{Contributions}
\begin{itemize}[leftmargin=*, itemsep=0.25em]
  \item \textbf{Agenthood in Six Birds terms.} We define an agent as a maintained package---a \emph{theory object}---whose interface variables make stable counterfactual differences at an induced scale (``an agent is a \emph{theory object}'').
  \item \textbf{Operational metrics for the six primitives.} We connect packaging (endomap + idempotence defect), viability (greatest fixed-point kernel), and difference-making (feasible empowerment) to the SBT primitives.
  \item \textbf{A minimal substrate with primitive toggles.} We provide a finite ring-world environment whose switches realize protocol holonomy, accounting/constraints, identity staging, and operator rewriting.
  \item \textbf{Evidence suite with guardrails.} We demonstrate (repair $\Rightarrow$ objecthood), (protocol $\Rightarrow$ horizon control capacity), (learning $\Rightarrow$ increased empowerment), and two null regimes that prevent false positives.
  \item \textbf{Reproducibility and formal anchor.} We ship hashed configs, an artifact auditor, and a Lean lemma establishing the finite greatest-fixed-point property underlying viability iteration.
\end{itemize}

\section{Dictionary: from six birds to agency}
\label{sec:dictionary}

\SBT\ (\citep{six_birds_theory}) treats macroscopic structure as \emph{induced} by six emergence primitives (``six birds''). Our agency proposal follows the same discipline: we do not assume goals, utilities, or homunculi; we assume only a finite stochastic substrate and ask which induced variables become stable, budgeted, and difference-making. The point of this section is to make the mapping explicit: each primitive contributes a distinct \emph{agency role}, and our experiments instrument that role with a corresponding metric or artifact.

\paragraph{Terminology: theory versus theory object.}
In \SBT\ a \emph{theory} is a layer/closure: an induced macro-physics $T=(\Pi,L,\mathcal F,B)$. Objects are then \emph{objects of a theory}: stable fixed points (or images) of the closures induced by $T$ under appropriate lenses and horizons. Throughout this paper we reserve the word ``theory'' for the induced layer $T$ and use \emph{theory object} for the stabilized package inside $T$. An \emph{agent} is a theory object with (i) a ledger-gated feasible interface alphabet and (ii) nontrivial within-layer difference-making.

\paragraph{Three operational faces of agenthood.}
In this paper we repeatedly return to three measurable proxies, each corresponding to a different aspect of the ``agent is a theory object'' thesis.
\begin{itemize}[leftmargin=*, itemsep=0.25em]
  \item \textbf{Packaging / objecthood:} does a macro lens that hides micro-degrees of freedom behave like an object map? We measure this with an empirical endomap $E$ and its idempotence defect (Section~\ref{sec:engine}, Exhibit~\ref{sec:ex_packaging}).
  \item \textbf{Budgeted viability:} does there exist a robust controlled-invariant safe set $\K$ under ledger-gated feasibility? We measure this with the viability kernel size $|\K|$ (Section~\ref{sec:engine}, Exhibit~\ref{sec:ex_packaging} and the sweep).
  \item \textbf{Difference-making:} does the interface admit more than one feasible policy that produces distinct outside futures? We measure this with feasible empowerment (channel capacity restricted by budgets and costs) (Section~\ref{sec:engine}, Exhibit~\ref{sec:ex_learning} and the holonomy plot).
\end{itemize}
These three proxies are deliberately orthogonal: packaging can exist without control, control without persistence is fragile, and persistence without budgets is not autonomous.

\begin{table}[t]
\centering
\small
\begin{tabular}{@{}p{0.13\linewidth} p{0.37\linewidth} p{0.42\linewidth}@{}}
\toprule
\textbf{Primitive} & \textbf{Agency role (interpretation)} & \textbf{Operationalization in this paper (metric / artifact)} \\
\midrule
P$_1$ Operator-write &
\emph{Skill / learning as induced law change.} The agent rewrites its effective transition operator so the same interface moves become more reliable or cheaper. &
A discrete skill variable $\theta$ reduces slip/noise in the ring-world dynamics. We measure increased difference-making as higher feasible empowerment at larger $\theta$ (learning--$\theta$ plot and medians). \\[0.35em]

P$_2$ Feasible-set-write &
\emph{Constraints define action.} ``Actions'' are moves inside an admissible set; infeasible interface commands collapse to no-ops. &
Ledger-gated feasibility: an action executes only if $r \ge \mathrm{cost}(a)$. We compute feasible empowerment by filtering sequences by budget and compute viability under feasible actions. Ablation: \texttt{constraints\_off}. \\[0.35em]

P$_3$ Protocol cycle (holonomy) &
\emph{Order matters.} Noncommuting interface moves create reachability that is invisible at one step but emerges over horizons. &
A phase-dependent displacement makes LEFT/RIGHT noncommutative across time when protocol is ON. Evidence: empowerment vs horizon $H$ shows a gap for $H\ge 2$ (holonomy plot). \\[0.35em]

P$_4$ Quantized identity / staging &
\emph{Robust tokens and horizons.} Discrete identity/staging variables support persistent reference frames and meaningful time scales. &
The environment includes discrete sectors (identity $g$ and phase $\phi$) that are invariant/staged. This supplies a stable tokenization of ``who'' and ``when''; it also explains why certain horizons cannot be idempotent when the lens includes $\phi$. \\[0.35em]

P$_5$ Order-theoretic closure &
\emph{Packaging as a maintained equivalence.} A closure induces macro variables that behave like objects under coarse lenses. &
Viability kernel $\K$ as a greatest fixed point under robust support semantics, plus packaging endomap $E$ whose idempotence defect quantifies objecthood for lenses that ignore microstate (packaging figure; Lean fixed-point lemma anchors the iteration). \\[0.35em]

P$_6$ Resource transduction / accounting &
\emph{Autonomy as paying for persistence.} Resources are stored and spent to maintain boundary and repair noise-induced damage. &
Ledger $r$ and maintenance action REPAIR convert budget into stability: repair collapses packaging defect and sustains viability under noise. Evidence: packaging ring (defect drop) and noise--maintenance sweep heatmaps. \\
\bottomrule
\end{tabular}
\caption{Dictionary from the six emergence primitives (\SBT) to agency roles and the corresponding operational metrics/artifacts used in this paper. Each row should be read as: ``if this primitive is absent or weakened, a specific aspect of agenthood/agency degrades in a characteristic way.''}
\label{tab:dictionary}
\end{table}

\paragraph{Guardrails (null regimes).}
Finally, we emphasize a methodological point already present in the \SBT\ program (\citep{six_birds_life}): induced notions are only meaningful if they survive obvious adversarial baselines. We therefore include two null regimes: (i) a single-action system yields zero empowerment for all horizons, and (ii) an exogenous schedule can fake empowerment if it is incorrectly treated as a controllable choice rather than as state/noise. These baselines are part of the evidence suite, not afterthoughts.

\section{The packaging engine: from kernels to induced agent variables}
\label{sec:engine}

This section fixes the computational substrate used throughout the paper and defines the operational proxies for the \SBT\ primitives. The guiding principle is the thesis from Section~\ref{sec:introduction}: \emph{agenthood is an enablement/closure claim, while agency is a within-layer causal claim}. We therefore define (i) what it means for an induced agent layer to exist (packaging and viability under budgets), and (ii) what it means for interface choices to be difference-making once such a layer exists (feasible empowerment).

\subsection{Typing: theories (layers) and theory objects}
A potential confusion is that in \SBT\ the word ``theory'' names the induced layer itself. We follow that usage. The finite data of this paper---a packaging lens, an induced controlled kernel, a feasibility gate, and a ledger/budget rule---constitute a theory (layer) $T$ at the chosen scale.

\paragraph{Definition (Theory / layer, specialized).}
In the notation of \SBT, a theory can be written $T=(\Pi,L,\mathcal F,B)$. In our finite setting, $\Pi$ is implemented by state projections, $L$ is the controlled kernel $P$, $\mathcal F$ is the ledger-gated feasible action set $A_{\mathrm{feas}}(s)$, and $B$ is the budget/ledger dynamics encoded in the state.

\paragraph{Definition (Theory object).}
A theory object is an object \emph{inside} a theory: a packaged macro label (or invariant package) that is stable under the closures induced by $T$ at the relevant horizon. Operationally, low idempotence defect means the macro labels behave like objects under composition; nonempty viability kernel means such objects can persist under robust support semantics. An \emph{agent} is a theory object equipped with an interface alphabet whose feasible interventions induce a nontrivial action channel to outside outputs (positive feasible empowerment).

\subsection{Microstate factoring and packaging}
We begin with a generic viewpoint that matches \SBT: a substrate microstate $x$ can be factored into \emph{inside}, \emph{boundary}, and \emph{outside} components,
\[
x \equiv (i,b,e),
\]
where ``boundary'' means the coupling degrees of freedom through which inside and outside exchange information and resources. A \emph{package} is an induced description (a coarse lens) $\Pi$ that produces a persistent macro-variable set. In later \SBT\ layers this could look like
\[
\Pi(i,b,e) \approx (s,o,r,y),
\]
where $s$ is an internal macrostate, $o$ is an observation at the boundary, $r$ is a ledger/resource variable, and $y$ is an outside macrostate. In this paper we do not assume geometry or rich observation models; instead we work in an exact finite-state setting where the package is implemented as projections of a finite state index.

\subsection{Finite controlled kernels}
\paragraph{Definition (Finite controlled kernel).}
Let $S=\{0,1,\dots,n-1\}$ be a finite state space and $A=\{0,1,\dots,m-1\}$ a finite action set. A controlled stochastic kernel is a row-stochastic tensor
\[
P[a,s,s'] \equiv \Pr(s_{t+1}=s' \mid s_t=s, a_t=a),
\]
with $P[a,s,s']\ge 0$ and $\sum_{s'}P[a,s,s']=1$ for all $(a,s)$. For a state distribution $d$ (row vector), one step under action $a$ is $d' = d\,P[a,\cdot,\cdot]$.

\paragraph{Definition (Successor support).}
We use robust support semantics: the set of possible successors under $(s,a)$ is
\[
\mathrm{Post}(s,a) \;:=\; \{\, s' \in S \;:\; P[a,s,s']>0 \,\}.
\]
All viability statements below are with respect to $\mathrm{Post}$ (not expectations): a move is safe only if \emph{every} nonzero-probability successor stays safe.

\subsection{Ledger-gated feasibility (constraints + accounting)}
\paragraph{Definition (Ledger and feasibility).}
A ledger is a nonnegative scalar extracted from state, $r:S\to \mathbb{R}_{\ge 0}$. Each action has a nonnegative cost $c:A\to \mathbb{R}_{\ge 0}$. The feasible actions at state $s$ are
\[
A_{\mathrm{feas}}(s) \;:=\; \{\, a\in A \;:\; c(a)\le r(s)\,\}.
\]
This is the operational form of P$_2$ (constraints) and P$_6$ (accounting): infeasible interface commands are not ``actions'' in the induced layer.

\subsection{Viability kernel as a greatest fixed point (P$_5$ backbone)}
\paragraph{Definition (Safe set).}
A safety predicate $\mathrm{Safe}:S\to\{\text{true},\text{false}\}$ specifies which states count as viable. We emphasize that ``safe'' can encode more than survival; in later sections we also use safety to encode coherence (e.g., requiring a repaired/noise-free internal bit).

\paragraph{Definition (Viability operator / controlled-invariance map).}
Given a candidate set $K\subseteq S$, define the contracting controlled-invariance operator
\[
\mathcal{V}(K) \;:=\; \Bigl\{\, s\in K \;:\; \mathrm{Safe}(s)\ \wedge\ \exists a\in A_{\mathrm{feas}}(s)\ \text{s.t.}\ \mathrm{Post}(s,a)\subseteq K \Bigr\}.
\]
Intuitively: $s$ remains in $K$ if it is safe and there exists a feasible action whose entire successor support stays inside $K$. The explicit condition $s\in K$ makes $\mathcal{V}$ pointwise contracting ($\mathcal{V}(K)\subseteq K$), matching the standard ``greatest controlled-invariant subset'' construction and the assumptions of our Lean fixed-point anchor.

\paragraph{Definition (Viability kernel).}
The viability kernel $\K\subseteq S$ is the greatest fixed point of $\mathcal{V}$ \citep{aubin1991,tarski1955},
\[
\K \;=\; \mathcal{V}(\K),
\]
computed by iterating $\mathcal{V}$ from the top safe set until convergence. In finite state spaces, this monotone decreasing iteration stabilizes in at most $|S|$ steps; our Lean lemma in the reproducibility section provides a formal anchor for this finite greatest-fixed-point property.

\paragraph{Remark (feedback versus open-loop).}
The viability kernel is a feedback notion: it certifies that for each state in $\K$ there exists an admissible action keeping successor support inside $\K$, and this choice can depend on the current state. Our empowerment calculations, by contrast, treat open-loop action sequences as channel inputs. We use these as complementary proxies: $\K$ certifies maintained existence under robust feasibility, while empowerment measures within-layer difference-making of the interface channel once such a viable domain exists.

Operationally, $|\K|$ serves as a quantitative proxy for ``agenthood as maintained existence'': if $\K=\emptyset$, no policy can keep the package viable under the ledger-gated constraints.

\subsection{From action sequences to channels}
\paragraph{Definition (Output lens).}
An output variable is a projection (lens) $f:S\to Y$ into a finite output set $Y=\{0,1,\dots,|Y|-1\}$. In the ring-world exhibits, $f$ is typically the outside position coordinate $y$ or a macro-lens $(y,r,\phi)$ that intentionally hides micro-variables.

\paragraph{Definition (Action-sequence channel).}
Fix an initial state $s_0$ and a horizon $H\ge 1$. For an action sequence $\alpha=(a_0,\dots,a_{H-1})$, let $S_H$ be the random state after rolling out the kernel under $\alpha$, and define $Y=f(S_H)$. The induced channel from sequences to outputs is
\[
W[\alpha, y] \;:=\; \Pr(Y=y \mid s_0,\alpha).
\]
Equivalently, each row of $W$ is the output distribution obtained by propagating $\delta_{s_0}$ through the sequence and aggregating probability mass by $f$.

\subsection{Feasible empowerment as difference-making}
\paragraph{Definition (Empowerment).}
For a fixed $(s_0,H,f)$, empowerment is the channel capacity of $W$ \citep{klyubin2005,shannon1948}:
\[
\Emp(s_0;H,f) \;:=\; \max_{p(\alpha)} I(\Alpha;Y),
\]
where $\Alpha$ is the chosen action-sequence random variable and $I(\cdot;\cdot)$ is mutual information \citep{shannon1948}. We compute this exactly (for small channels) via Blahut--Arimoto \citep{blahut1972,arimoto1972}.

\paragraph{Definition (Feasible empowerment).}
To respect P$_2$/P$_6$, we restrict the sequence set by ledger feasibility. Given costs $c(a)$ and budget $r(s_0)$, a sequence $\alpha$ is feasible if $\sum_{t=0}^{H-1} c(a_t) \le r(s_0)$. Feasible empowerment is the capacity of the restricted channel:
\[
\Emp_{\mathrm{feas}}(s_0;H,f) \;:=\; \mathrm{Cap}\bigl(W\ \text{restricted to feasible }\alpha\bigr).
\]
\paragraph{Remark (what ``feasible'' means in this paper).}
In principle, feasibility can be state- and branch-dependent because the ledger is part of the state and may evolve over the horizon. In this paper we use a deliberately simple, explicit notion: we treat the channel input alphabet as \emph{open-loop} action sequences and restrict it by an \emph{initial-budget} gate, i.e.\ $\sum_{t=0}^{H-1} c(a_t)\le r(s_0)$. This yields an ``open-loop budgeted empowerment'' that is easy to audit and compare under matched controls. It can over- or under-approximate stricter stepwise feasibility in settings with replenishment or branch-dependent constraints; we therefore treat it as an operational proxy and include calibrated nulls (Section~\ref{sec:ex_nulls}) to guard against mis-typed control channels.

\paragraph{Aggregation.}
When we report ``median empowerment on $\K$,'' we compute $\Emp_{\mathrm{feas}}(s;H,f)$ per state $s\in\K$ and take the median; if $\K$ is large, we use a fixed deterministic subset rule recorded in the artifact manifest to keep computation bounded while remaining reproducible.
When $\Emp_{\mathrm{feas}}\approx 0$ (and null regimes confirm it is not an artifact), the induced layer has no nontrivial action channel: it cannot ``throw a stone'' in the counterfactual sense.

\subsection{Packaging endomap and idempotence defect (objecthood proxy)}
Packaging is not only about survival; it is about whether a coarse lens induces stable object-like labels when micro-degrees of freedom churn underneath.

\paragraph{Definition (Macro lens and fibers).}
Let $\pi:S\to X$ be a macro lens (coarser than $f$ in general). For a macro label $x\in X$, define its fiber $S_x := \{s\in S : \pi(s)=x\}$.

\paragraph{Definition (Empirical packaging endomap).}
Fix a horizon $\tau\ge 0$ and a stationary policy $\mu(a\mid s)$ (deterministic or stochastic). For each macro label $x$, initialize a reference micro-distribution $d_0$ that is uniform over $S_x$. Let $T_\mu$ be the induced one-step transition under $\mu$:
\[
T_\mu[s,s'] := \sum_{a\in A} \mu(a\mid s)\,P[a,s,s'].
\]
Evolve $d_\tau := d_0\,T_\mu^\tau$, aggregate to a macro distribution over $X$ via $\pi$, and define $E(x)$ as the mode macro label at time $\tau$ (ties broken deterministically by smallest label). The resulting map
\[
E: X\to X
\]
is an empirical \emph{packaging endomap}: it says ``starting from macro label $x$ (with hidden microstate randomized), what macro label do we typically return to after $\tau$ steps under the policy?''

\paragraph{Definition (Idempotence defect).}
The idempotence defect of $E$ is
\[
\mathrm{Def}(E) \;:=\; \frac{|\{x\in X: E(E(x))\neq E(x)\}|}{|X|}.
\]
If $\mathrm{Def}(E)\approx 0$, the macro labels behave like stable objects under the lens $\pi$ and the policy $\mu$; if $\mathrm{Def}(E)$ is large, the lens fails to induce objecthood (labels do not compose stably).

\subsection{Relativity to lenses, horizons, and maintenance policies}
All operational notions in this paper are relative to a chosen induced layer (a theory in \SBT\ terms) and to explicit observational/temporal choices: (i) what counts as ``outside'' (output lens $f$), (ii) what macro labels are treated as objects (macro lens $\pi$), (iii) horizons $H$ (for difference-making) and $\tau$ (for packaging stability), and (iv) a maintenance policy $\mu$ used to define the packaging endomap $E$. This is not an arbitrary observer effect layered on top of the model: in \SBT\ the lenses and horizons are part of specifying the layer itself, and our claims are comparative under matched choices of $(f,\pi,H,\tau,\mu)$.

\paragraph{Remark (policy dependence is intended).}
The endomap $E$ is conditioned on a policy $\mu$ because many objects are maintained rather than passive. We therefore interpret low idempotence defect as a statement about objecthood \emph{under a maintenance regime}, not as a claim that the macro labels are self-stabilizing under all policies.

\paragraph{Definition (Agent as a theory object, operational).}
Fix a theory layer $T$ (kernel + feasibility + ledger), lenses $(f,\pi)$, horizons $(H,\tau)$, and a maintenance policy $\mu$ for $E$. We call the induced package an \emph{agent theory object at this scale} if (i) the viability kernel $\K$ is nonempty for the chosen safety predicate (maintained existence), (ii) the macro labels behave object-like under $\pi$ and $\mu$ at horizon $\tau$ (low idempotence defect), and (iii) the within-layer action channel to $f$ at horizon $H$ is nontrivial (positive feasible empowerment on $\K$). The exhibits report these as $|\K|$, median empowerment on $\K$, and $\mathrm{Def}(E)$.

\subsection{Claims versus evidence (mini-map)}
\begin{center}
\fbox{\begin{minipage}{0.94\linewidth}
\small
\textbf{Claims vs evidence.} Each primitive contributes a distinct agency role; we test them with complementary artifacts.
\begin{itemize}[leftmargin=*, itemsep=0.2em]
  \item \textbf{Packaging requires maintenance (P$_5$+P$_6$).} Repair collapses idempotence defect under a lens that hides microstate (Exhibit~\ref{sec:ex_packaging}); the noise--maintenance sweep shows where viability and empowerment collapse when repair becomes unaffordable (Exhibit~\ref{sec:ex_sweep}).
  \item \textbf{Difference-making is not automatic.} Single-action systems have zero empowerment, and exogenous schedules can fake empowerment if mis-modeled (Exhibit~\ref{sec:ex_nulls}).
  \item \textbf{Protocol holonomy creates horizon-dependent control (P$_3$).} Empowerment separates protocol ON vs OFF only at $H\ge 2$ (Exhibit~\ref{sec:ex_holonomy}).
  \item \textbf{Operator rewriting thickens causal control (P$_1$).} Increasing skill $\theta$ monotonically increases empowerment by reducing effective noise (Exhibit~\ref{sec:ex_learning}).
  \item \textbf{Ablations connect the pieces (P$_1$--P$_6$).} A small toggle suite summarizes how the metrics change when primitives are weakened (Exhibit~\ref{sec:ex_ablations}).
\end{itemize}
\end{minipage}}
\end{center}

\section{Exhibit: repair makes objecthood}
\label{sec:ex_packaging}

This exhibit isolates a minimal claim: \emph{packaging stability depends on maintenance}. In the engine (Section~\ref{sec:engine}) we defined an empirical packaging endomap $E$ for a macro lens $\pi$ that hides microstate, and an idempotence defect $\mathrm{Def}(E)$ that quantifies whether macro labels behave like stable objects under that lens. Here we show that allowing a repair/maintenance action (P$_6$) collapses defect from maximal to zero at a coherence-aligned horizon.

\subsection{Setup: a lens that hides microstate}
We use the ring-world substrate whose full microstate includes an outside coordinate $y$ on a ring, an internal damage bit $u$, a staged phase variable $\phi$, a ledger $r$, and (optionally) identity/skill sectors. The crucial choice is the macro lens:
\[
\pi(y,u,\phi,r,\dots)\;:=\;(y,r,\phi),
\]
i.e., the lens \emph{intentionally ignores} the damage bit $u$. This is exactly the situation in which ``objecthood'' is nontrivial: if $u$ churns underneath but the induced macro labels remain stable, then the lens has packaged the micro-churn into an object-like macro description; if the labels fail to compose, then the lens does not induce an object.

Because the lens includes the staged phase $\phi$ with period $m_{\phi}=2$, idempotence is only meaningful at horizons $\tau$ that respect staging. We therefore emphasize $\tau=2$ (a full phase cycle), which factors out the trivial non-idempotence that would occur at odd $\tau$ purely from the $\phi$ shift. Figure~\ref{fig:packaging_ring} reports $\mathrm{Def}(E)$ across $\tau$; we highlight $\tau=2$ because it is the \emph{first} staging-aligned horizon (a full phase cycle), i.e., the earliest point at which idempotence can hold without being blocked by the $\phi$ shift.

\subsection{Two regimes: repair disabled versus repair enabled}
We compare two regimes that are identical except for whether REPAIR is present and used.
\begin{itemize}[leftmargin=*, itemsep=0.25em]
  \item \textbf{Repair OFF:} the policy always moves RIGHT; the damage bit $u$ can flip due to noise, but the agent cannot pay to reset it.
  \item \textbf{Repair ON:} the policy uses REPAIR whenever $u=1$ (when feasible) and otherwise moves RIGHT; repair succeeds with probability $1$ when executed.
\end{itemize}
For each macro label $x=(y,r,\phi)$, we initialize the reference distribution uniformly over its fiber $S_x=\{s:\pi(s)=x\}$ and roll forward $\tau$ steps under the policy to define the endomap $E(x)$ as the mode macro label (Section~\ref{sec:engine}). The idempotence defect $\mathrm{Def}(E)$ then reports what fraction of macro labels fail the compositional stability test $E(E(x))=E(x)$.

\subsection{Result: defect collapses at $\tau=2$}
\begin{figure}[t]
  \centering
  \includegraphics[width=0.88\linewidth]{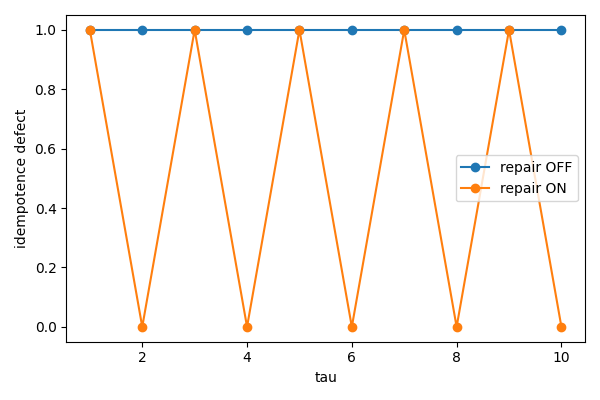}
  \caption{Packaging stability requires maintenance. Idempotence defect $\mathrm{Def}(E)$ of the empirical endomap $E$ under the macro lens $\pi(y,u,\phi,r,\dots)=(y,r,\phi)$ that hides the damage bit $u$. Repair disabled (policy cannot reset $u$) yields maximal defect at $\tau=2$; repair enabled and used collapses defect to zero at $\tau=2$.}
  \label{fig:packaging_ring}
\end{figure}

Figure~\ref{fig:packaging_ring} plots $\mathrm{Def}(E)$ versus horizon $\tau$ for the two regimes. At $\tau=2$ we obtain:
\[
\mathrm{Def}(E)_{\text{repair OFF}} = 1.0,
\qquad
\mathrm{Def}(E)_{\text{repair ON}} = 0.0.
\]
That is: without maintenance, the macro labels $(y,r,\phi)$ behave maximally non-object-like under composition; with maintenance, the same coarse lens becomes perfectly idempotent at the staging-aligned horizon.

\subsection{Interpretation in Six Birds terms}
This is the minimal ``objecthood requires payment'' statement. The lens $\pi$ is not enough by itself: if microstate can drift underneath with no compensating closure, then macro labels fail to behave like objects. Allowing repair introduces a P$_6$ transduction (spending budget to restore coherence) that, together with P$_5$ closure, stabilizes the induced macro description. In the language of the thesis, repair makes the induced \emph{theory layer} support stable \emph{theory objects}: at the staging-aligned horizon, the coarse macro labels behave like objects under composition, providing a well-typed substrate on which an agent object with an interface can be defined.

In later exhibits we show that this stabilized package also supports nontrivial difference-making (feasible empowerment) and survives explicit guardrails that would otherwise produce false positives.

\section{Exhibit: null regimes and trap checks}
\label{sec:ex_nulls}

The engine in Section~\ref{sec:engine} defines agency as \emph{difference-making}: there exist at least two feasible interface policies whose induced channel to outside futures has nonzero capacity. This is a powerful operational notion---and precisely because it is powerful, it is vulnerable to two classes of false positives:
(i) confusing motion with control, and
(ii) confusing exogenous structure with choice.

This exhibit provides two guardrails (null regimes) that every later empowerment claim must survive.

\subsection{Null A: single-action regimes have zero empowerment}
If there is only one action available everywhere, then there is no choice variable and hence no action channel. In the notation of Section~\ref{sec:engine}, the action-sequence channel $W[\alpha,y]$ has exactly one input row (one possible $\alpha$), so its capacity is identically zero for any horizon $H$.

We confirm this in an exact finite kernel with a nontrivial state evolution but a single action: the dynamics can move, but the agent cannot steer. The feasible empowerment values (bits) are:
\[
\Emp_{\mathrm{feas}}(H=1)=0,\quad
\Emp_{\mathrm{feas}}(H=2)=0,\quad
\Emp_{\mathrm{feas}}(H=3)=0.
\]
This baseline prevents an error of interpretation: \emph{being a dynamical subsystem is not the same as having agency}. Agency requires an input degree of freedom that can be intervened upon.

\subsection{Null B: the schedule trap (exogenous structure mis-modeled as choice)}
A more subtle failure mode is the \emph{schedule trap}. Many environments contain an exogenous variable---a clock, schedule, or external forcing---that influences the next outside state. If we mistakenly treat this exogenous variable as if it were the agent's action, we manufacture a fictitious control channel and obtain spurious empowerment.

We demonstrate this with a toy system where an exogenous schedule bit $s_{\mathrm{ext}}$ determines the next outside state $x$. In the \emph{correct} model, $s_{\mathrm{ext}}$ is part of the state/noise and the agent's actions are ineffective; the induced channel rows are identical and capacity is zero. In the \emph{incorrect} model, we collapse state and treat the schedule as a controllable ``action'' that directly sets $x$; this produces an apparent 1-bit channel even though no such control exists.

\begin{table}[t]
\centering
\small
\begin{tabular}{@{}l l@{}}
\toprule
\textbf{Null regime} & \textbf{Empowerment / capacity (bits)} \\
\midrule
Null A (single action), $H=1$ & $0.0$ \\
Null A (single action), $H=2$ & $0.0$ \\
Null A (single action), $H=3$ & $0.0$ \\
\addlinespace
Null B (schedule trap), \emph{wrong} model & $1.0$ \\
Null B (schedule trap), \emph{right} model & $0.0$ \\
\bottomrule
\end{tabular}
\caption{Guardrails against false agency. Null A shows that dynamics without choice yields zero empowerment at all horizons. Null B shows that exogenous schedules can fake empowerment if incorrectly modeled as controllable actions; treating the schedule as state/noise restores the correct capacity of zero.}
\label{tab:nulls}
\end{table}

\subsection{Why these nulls matter for the thesis}
The thesis ``an agent is a \emph{theory object}'' includes a modeling discipline: the underlying theory/layer must type variables correctly on the proper side of the interface. Null A prevents us from calling any persistent moving pattern an agent. Null B prevents us from smuggling outside structure into the agent by mislabeling it as action. The remaining exhibits therefore interpret empowerment only after (i) feasibility is enforced by the ledger (P$_2$/P$_6$), and (ii) the action variable is genuinely controllable at the induced scale.

\section{Exhibit: protocol holonomy creates horizon-dependent control}
\label{sec:ex_holonomy}

This exhibit isolates P$_3$ (protocol cycle / holonomy): \emph{order matters}. In the engine (Section~\ref{sec:engine}) we defined (feasible) empowerment as the capacity of the induced channel from action sequences to outside futures. P$_3$ predicts a distinctive signature: if the environment admits noncommuting interface moves, then one-step control can look identical across regimes, while multi-step horizons reveal additional controllability created purely by \emph{sequence structure}.

\subsection{Setup: identical kernels except for protocol}
We compare two ring-world regimes that are identical in noise, budgets, and costs; the only difference is whether protocol holonomy is enabled.
\begin{itemize}[leftmargin=*, itemsep=0.25em]
  \item \textbf{Protocol ON (P$_3$ enabled):} the displacement induced by LEFT/RIGHT depends on a staged phase variable $\phi$, so that action composition over time is noncommutative (the effective move depends on when it is applied).
  \item \textbf{Protocol OFF:} LEFT/RIGHT are phase-independent; sequence order produces no additional geometric drift beyond the obvious one-step effects.
\end{itemize}
In both regimes we measure \emph{median feasible empowerment on the viability kernel} $\K$ using output lens $f(s)=y$ (outside ring position) and horizons $H\in\{1,2,3,4,5\}$. Feasibility is enforced by the ledger: an action sequence is admissible only if its total cost fits within the initial budget (Section~\ref{sec:engine}).

\subsection{Result: equality at $H=1$, divergence for $H\ge 2$}
\begin{figure}[t]
  \centering
  \includegraphics[width=0.88\linewidth]{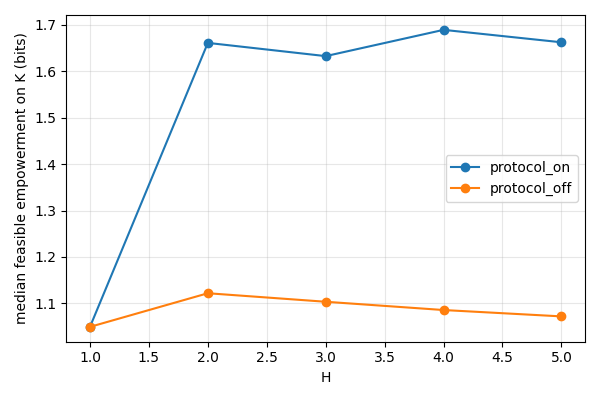}
  \caption{Protocol holonomy yields horizon-dependent control. Median feasible empowerment (bits) on the viability kernel $\K$ as a function of horizon $H$ for protocol ON vs OFF, using output lens $f(s)=y$ (outside position). The curves coincide at $H=1$ but diverge for $H\ge 2$, as predicted by P$_3$ noncommutativity: ordered action composition creates additional reachable outside futures that are not present in the phase-independent regime.}
  \label{fig:protocol_horizon}
\end{figure}

Figure~\ref{fig:protocol_horizon} shows the median feasible empowerment (bits) as a function of horizon $H$.

\emph{At $H=1$, the curves coincide.} This is expected: a single step cannot exploit protocol holonomy because there is no order-of-composition effect in a one-element sequence. Formally, the channel inputs are single actions, so any holonomy that arises from noncommuting compositions is invisible.

\emph{For $H\ge 2$, the curves separate.} Once sequences have length at least two, noncommutativity can create distinct reachable output distributions that do not exist in the protocol-OFF regime. This is the operational content of P$_3$: agency can arise from the geometry of action composition even when the one-step menu of actions looks the same.

For reference, the measured empowerment values (median over viable states) are:

\[
\Emp_{\mathrm{feas}}^{\text{protocol ON}}(H=1..5)
=
[1.0493,\ 1.6613,\ 1.6328,\ 1.6894,\ 1.6627],
\]
\[
\Emp_{\mathrm{feas}}^{\text{protocol OFF}}(H=1..5)
=
[1.0493,\ 1.1218,\ 1.1034,\ 1.0856,\ 1.0719].
\]

\subsection{A checkable noncommutativity witness}
Beyond the aggregate curve, we exhibit a single start state $s^\star$ and two length-2 sequences $\alpha=(\mathrm{R},\mathrm{L})$ and $\beta=(\mathrm{L},\mathrm{R})$. In the protocol-ON kernel their induced output distributions over $y$ differ, while in the protocol-OFF kernel they coincide (up to numerical tolerance). We report this as a total variation distance:
\[
\mathrm{TV}\!\left(W_{\alpha},W_{\beta}\right)
=
\begin{cases}
0.6720 & \text{protocol ON},\\
0.0960 & \text{protocol OFF},
\end{cases}
\qquad
s^\star:\ (y=0,\ r=2,\ \phi=1,\ u=0),\ 
\alpha=(\mathrm{R},\mathrm{L}),\ \beta=(\mathrm{L},\mathrm{R}).
\]

This is the concrete sense in which order-of-composition creates additional distinguishable futures at $H\ge 2$ even when $H=1$ remains matched.

\subsection{Interpretation in Six Birds terms}
This is the simplest ``protocol makes an agent more than a thermostat'' result. P$_3$ is not memory (P$_5$) and not budget (P$_6$): it is purely geometric. It says that a controller can gain effective degrees of freedom from the \emph{order} of its interface moves, and that such gains only appear when the induced layer admits multi-step horizons. In the thesis language, protocol holonomy enriches the agent object's action channel within an otherwise fixed theory: it increases the expressivity of feasible interventions without changing the packaging lens, thickening what the interface can reliably cause.

\section{Exhibit: primitive ablations (a quantitative toggle table)}
\label{sec:ex_ablations}

The previous exhibits isolated single mechanisms (maintenance for packaging; protocol for horizon-dependent control) and validated null baselines. This exhibit summarizes the same story as a compact toggle suite: we weaken or disable specific mechanisms corresponding to the \SBT\ primitives and measure the three operational proxies defined in Section~\ref{sec:engine}:
(i) viability kernel size $|\K|$,
(ii) median feasible empowerment on $\K$ (bits), and
(iii) packaging idempotence defect for a coarse macro lens.

\subsection{Ablation suite and metrics}
Each row of Table~\ref{tab:ablations} corresponds to a configuration of the ring-world environment with a small number of toggles (protocol on/off, repair on/off, constraint costs on/off, learning on/off, noise strength, repair success). The table is generated from audited artifacts (Section~\ref{sec:repro}) and is included here verbatim to keep the paper aligned with reproducible outputs.

\begin{table}[t]
\centering
\small
\begin{tabular}{lrrr}
\toprule
name & kernel size viable & empowerment median on K & idempotence defect \\
\midrule
constraints off & 64 & 0.794 & 1.000 \\
full & 64 & 1.661 & 0.333 \\
high noise & 64 & 1.153 & 0.333 \\
learn on & 128 & 1.831 & 0.333 \\
no protocol & 64 & 1.122 & 0.333 \\
no repair & 0 & 0.000 & 0.000 \\
repair imperfect & 64 & 1.557 & 0.000 \\
\bottomrule
\end{tabular}

\caption{Primitive ablations summary (generated). Columns: viability kernel size $|\K|$, median feasible empowerment on $\K$ (bits) at horizon $H=2$ using output lens $f(s)=y$, and packaging idempotence defect (Section~\ref{sec:engine}). Higher $|\K|$ indicates a larger robust controlled-invariant safe set under ledger-gated feasibility; higher empowerment indicates greater counterfactual difference-making; lower defect indicates more object-like macro labels under a coarse lens. Unless otherwise stated in an exhibit, viability here uses a ledger-only safety predicate $\mathrm{Safe}(s):=(r(s)\ge 1)$ (the sweep exhibit additionally requires $u=0$ to treat coherence as safety). All rows use the same observational contract for comparability: output lens $f(s)=y$, packaging lens $\pi=(y,r,\phi)$, empowerment horizon $H=2$, and packaging horizon $\tau=2$ (unless an exhibit explicitly states a different safety predicate or lens).}
\label{tab:ablations}
\end{table}

\subsection{Reading the table in Six Birds terms}
The table makes three structural points explicit.

First, \textbf{agenthood is not guaranteed by dynamics alone}. When maintenance/repair is removed under costs (the \texttt{no repair} row), the viability kernel collapses ($|\K|=0$). In the engine semantics, this means there exists no state from which any feasible policy can keep all successor support inside a safe set. If the package cannot be maintained, there is no stable agent layer in which ``actions'' persist.

Second, \textbf{difference-making depends on protocol structure, not only on action availability}. Comparing \texttt{full} to \texttt{no protocol}, we see the same $|\K|$ and the same packaging defect, but a substantial drop in feasible empowerment when protocol holonomy is disabled. This matches the P$_3$ claim from Exhibit~\ref{sec:ex_holonomy}: noncommutativity enriches the reachable outside futures only at multi-step horizons.

Third, \textbf{constraints and accounting shape what counts as an action channel}. The \texttt{constraints off} row illustrates that removing costs changes which sequences are feasible and therefore changes both empowerment and the viability story---but it does not automatically produce objecthood. Packaging stability is a separate axis (captured by idempotence defect) that depends on closure and maintenance, not on ``freedom to act'' alone.

The remaining rows (e.g., increased noise, imperfect repair, and learning/skill) should be read as quantitative perturbations: weakening the maintenance channel or increasing noise shrinks viable structure and reduces difference-making, while operator rewriting can recover some control capacity by reducing effective noise (expanded in Exhibit~\ref{sec:ex_learning}).

\subsection*{Takeaways}
\begin{itemize}[leftmargin=*, itemsep=0.25em]
  \item \textbf{Maintenance gates existence.} Without budgeted maintenance, the robust viability kernel can collapse to $\emptyset$, leaving no stable agent layer.
  \item \textbf{Protocol creates control beyond one step.} With the same viability and packaging, enabling holonomy increases feasible empowerment by expanding horizon-dependent reachability.
  \item \textbf{Action is feasibility, not fantasy.} Changing costs and admissibility reshapes empowerment and viability, but objecthood (packaging defect) remains a distinct requirement tied to closure and repair.
\end{itemize}

\section{Exhibit: noise--maintenance sweep (a phase diagram)}
\label{sec:ex_sweep}

The packaging exhibit (Section~\ref{sec:ex_packaging}) showed that maintenance can collapse idempotence defect for a coarse lens that hides microstate. Here we probe the complementary question: \emph{when does the agent layer exist at all under noise?} In the engine (Section~\ref{sec:engine}) we defined viability as a robust, support-based greatest fixed point. This exhibit shows a phase-diagram style transition: as internal noise increases, the viability kernel and the difference-making capacity collapse unless repair is cheap enough to be routinely feasible.

\subsection{Setup: sweeping noise and maintenance cost}
We use the ring-world environment with a damage bit $u$ that can flip due to noise and a repair action REPAIR that (when executed) restores $u=0$. We sweep a modest $8\times 8$ grid over:
\begin{itemize}[leftmargin=*, itemsep=0.25em]
  \item \textbf{Noise:} $p_{\mathrm{flip}}$ (probability of damage), from $0$ to $0.7$.
  \item \textbf{Maintenance cost:} $\mathrm{cost}(\text{REPAIR})$ from $0$ to $7$.
\end{itemize}
All other parameters are held fixed (ring size, phase setting, and a ledger model with periodic gains), and repair succeeds with probability $1$ when it is feasible.

To make the viability notion sensitive to coherence (not merely survival), we choose a safe predicate that requires both budget and repaired state:
\[
\mathrm{Safe}(s) \;:=\; (r(s)\ge 1)\ \wedge\ (u(s)=0).
\]
This is a deliberate modeling choice: it treats the unrepaired internal bit as a loss of coherent objecthood, so the viability kernel measures ``staying viable \emph{and} coherent'' under support semantics.

\subsection{Measured quantities}
At each grid point we compute two quantities from Section~\ref{sec:engine}:
\begin{enumerate}[leftmargin=*, itemsep=0.25em]
  \item \textbf{Viability kernel size} $|\K|$ under ledger-gated feasibility and successor-support invariance.
  \item \textbf{Median feasible empowerment on $\K$} (bits) at horizon $H=2$ using output lens $f(s)=y$ (outside ring position), with sequences filtered by the initial budget.
\end{enumerate}
If $\K=\emptyset$, empowerment is defined as $0$ by convention (no viable states, hence no induced agent layer).

\subsection{Result: a collapse boundary in $(p_{\mathrm{flip}}, \mathrm{cost})$ space}
\begin{figure}[t]
  \centering
  \includegraphics[width=0.92\linewidth]{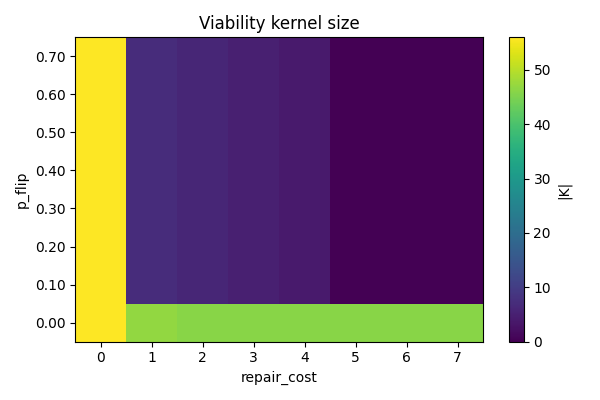}

  \vspace{0.7em}

  \includegraphics[width=0.92\linewidth]{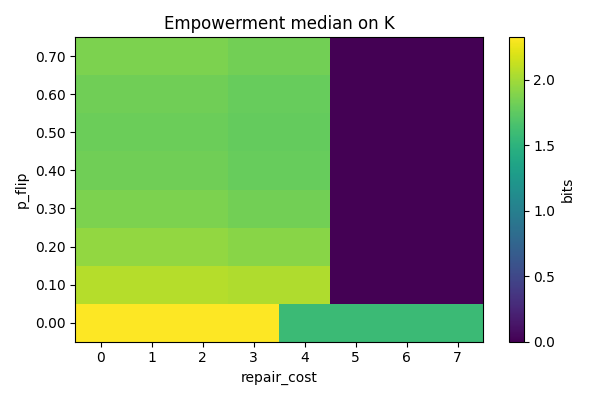}
  \caption{Noise--maintenance sweep (phase diagram). \textbf{Top:} viability kernel size $|\K|$ under the safe predicate $(r\ge 1)\wedge(u=0)$ and robust successor-support semantics. \textbf{Bottom:} median feasible empowerment (bits) on $\K$ at horizon $H=2$ with output lens $f(s)=y$. As noise increases and repair becomes expensive, the feasible set shrinks and the robust kernel collapses; empowerment collapses along the same boundary because difference-making requires an induced layer that can be maintained.}
  \label{fig:sweep_heatmaps}
\end{figure}

Figure~\ref{fig:sweep_heatmaps} shows the two heatmaps. The qualitative story is simple:
\begin{itemize}[leftmargin=*, itemsep=0.25em]
  \item Increasing noise $p_{\mathrm{flip}}$ makes it harder to keep $u=0$ in the robust-support sense: all nonzero-probability successors must remain safe.
  \item Increasing repair cost shrinks the feasible action set $A_{\mathrm{feas}}(s)$ (Section~\ref{sec:engine}), eventually making it impossible to guarantee repair when damage occurs.
\end{itemize}
Because we use successor-support (almost-sure under the modeled kernel) semantics, $\K$ can be nonempty under noise only when there exists at least one feasible action whose transition support avoids unsafe states (e.g., repair actions that restore $u=0$ with probability 1 in the model); if all actions admit any unsafe successor with nonzero probability, the robust kernel collapses by design.
The combined effect produces a region where $|\K|$ collapses to $0$ (no policy can keep the system safe-and-coherent for all support outcomes), and a complementary region where $|\K|$ remains substantial. Empowerment collapses along the same boundary because a nontrivial action channel requires an induced layer that persists: without viability, there is no stable domain on which actions are meaningful.

Quantitatively, over this grid we observe:
\[
|\K| \in [0.0,\ 56.0],
\qquad
\text{median }\Emp_{\mathrm{feas}} \in [0.0,\ 2.321928094887362]\ \text{bits}.
\]
(The empowerment maximum is approximately $\log_2 5 \approx 2.322$ bits in this configuration.)

\subsection{Interpretation in Six Birds terms}
This is the most direct ``agency requires maintenance under noise'' result. P$_6$ (accounting/transduction) supplies the mechanism for paying to restore coherence; P$_2$ (constraints) determines whether that mechanism is actually feasible at the boundary; and P$_5$ (closure) turns these into a robust invariant safe set under support semantics. The phase diagram makes the interaction visible: when the inequality implied by the ledger can no longer fund repair at the rate demanded by noise, the induced agent layer disappears ($\K=\emptyset$), and with it the meaning of action-as-difference-making.

\section{Exhibit: operator rewriting thickens causal control (learning $\theta$)}
\label{sec:ex_learning}

This exhibit isolates P$_1$ (operator-write): \emph{learning as induced law change}. In \SBT, P$_1$ is not merely storing facts; it is rewriting the effective transition operator so the same boundary interventions become more reliable, cheaper, or higher-bandwidth. In the engine (Section~\ref{sec:engine}) this should show up as increased difference-making capacity: higher feasible empowerment under comparable constraints.

\subsection{Setup: a discrete skill variable that rewrites dynamics}
We use a ring-world configuration with a discrete skill variable $\theta\in\{0,1,2\}$ that reduces effective slip/noise in the dynamics. Intuitively: higher $\theta$ corresponds to a better internal model or controller that executes the same LEFT/RIGHT interface moves with fewer unintended outcomes. This is the operational surrogate of P$_1$: changing $\theta$ changes the induced kernel $P[a,s,s']$ (the law), not merely the agent's memory.

To isolate the operator effect from budgets, we set action costs to zero in this configuration; feasibility is therefore trivial (all sequences are feasible), and empowerment reflects reliability/controllability rather than spending power. Since all action costs are zero in this configuration, feasibility is trivial and $\Emp_{\mathrm{feas}}=\Emp$ here. We measure empowerment at horizon $H=2$ using the outside position lens $f(s)=y$, and we summarize by the \emph{median} empowerment over viable states in each $\theta$-sector (restricting to a fixed staged phase and a coherent internal bit to avoid conflating $\theta$ with trivial staging effects).

\subsection{Result: empowerment increases monotonically with skill}
\begin{figure}[t]
  \centering
  \includegraphics[width=0.78\linewidth]{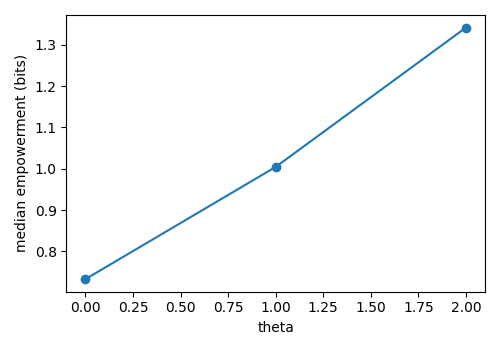}
  \caption{Operator rewriting (P$_1$) increases difference-making. Median feasible empowerment (bits) at horizon $H=2$ using output lens $f(s)=y$ (outside position), grouped by discrete skill $\theta$ that reduces effective slip/noise in the ring-world kernel. Empowerment increases monotonically with $\theta$, consistent with P$_1$ as an induced law change rather than merely an internal record.}
  \label{fig:learning_theta}
\end{figure}

Figure~\ref{fig:learning_theta} shows the median empowerment as a function of $\theta$. The measured medians (bits) are:
\begin{align*}
\mathrm{median}\ \Emp_{\mathrm{feas}}(\theta=0) &= 0.7330983751920465,\\
\mathrm{median}\ \Emp_{\mathrm{feas}}(\theta=1) &= 1.0045739033658136,\\
\mathrm{median}\ \Emp_{\mathrm{feas}}(\theta=2) &= 1.3416542136663907.
\end{align*}
Thus, higher skill yields a strictly larger action channel to outside futures at the same horizon.

\subsection{Interpretation: learning as ``causal thickening''}
This is the minimal learning-as-agency statement in \SBT terms. Packaging and viability (P$_5$/P$_6$) can make an induced layer persist, and protocol holonomy (P$_3$) can add horizon-dependent reachability, but P$_1$ changes something deeper: it changes the \emph{effective physics} of the induced layer. Increasing $\theta$ rewrites the transition structure so that interface interventions map to outside consequences with higher reliability; the induced theory becomes sharper, and the agent becomes more capable of throwing a stone in the counterfactual sense.

In the thesis language, this is the point at which ``agent as theory object'' becomes quantitative: operator rewriting (higher $\theta$) does not merely store information, it changes the effective transition law so the same interface interventions map to outside consequences with higher reliability. The theory remains the same induced layer, but the agent object becomes a better compiler of causes at the boundary, producing a higher-capacity action channel to outside futures.

\section{Reproducibility and artifact contract}
\label{sec:repro}

This paper is designed to be reproducible as a \emph{contract}, not as an aspiration. Every quantitative claim in the exhibits is backed by an audited artifact produced by deterministic scripts, keyed by stable configuration hashes. This section specifies (i) the contract that artifacts must satisfy, (ii) the exact commands that regenerate and verify them, and (iii) the small formal anchor that connects our viability computation to a greatest-fixed-point theorem.

\subsection{Artifact contract (what every result must contain)}
All experiment outputs live under \texttt{results/} and are machine-checkable. JSON artifacts are required to include (possibly with type-specific naming for multi-config comparisons):
\begin{itemize}[leftmargin=*, itemsep=0.25em]
  \item \textbf{Configuration identity:} a \texttt{config} dictionary and its \texttt{config\_hash}, where the hash is a stable JSON hash of the config.
  \item \textbf{Metrics payload:} a \texttt{metrics} dictionary (or a type-specific payload for traces/sweeps) containing the quantitative outputs used in the paper.
  \item \textbf{Provenance:} a timestamp field (e.g., \texttt{created\_at\_utc}) and a \texttt{versions} dictionary (Python, NumPy, Matplotlib as applicable).
\end{itemize}
Filenames also encode hashes when appropriate, so that a plot or table can be traced to a configuration without opening the file.

For paper-facing stability, we export stable-named assets (figures and generated tables) into \texttt{paper/figures/} and \texttt{paper/generated/}. The file \texttt{paper/generated/numbers.json} collects the key scalars cited in the exhibits (e.g., idempotence defect at $\tau=2$, empowerment arrays, and learning medians) together with the hashes of the runs they were extracted from.

\subsection{How to regenerate and verify (exact commands)}
To regenerate all artifacts used by the paper from scratch, and then verify them, run the following two commands from the repository root:

\begin{verbatim}
python scripts/run_all_experiments.py --clean
python scripts/audit_results.py --strict
\end{verbatim}

The first command rebuilds the full suite of evidence artifacts (rollouts, packaging measurements, null regimes, holonomy horizons, ablation table, sweep heatmaps, and learning--$\theta$ medians). The second command enforces the artifact contract above and fails if any artifact is missing required fields, if any \texttt{config\_hash} disagrees with the stable hash of the stored config, or if any stored probability objects violate basic stochasticity invariants.

\subsection{Determinism and traceability}
Experiments that involve sampling (e.g., sampling states from a viability kernel to estimate a median empowerment) use an explicit global seed and deterministic sampling rules. Configuration hashing uses a stable JSON serialization to ensure that the same config yields the same hash across runs and machines (modulo Python/library versions, which are recorded in each artifact).

\subsection{Formal anchor: viability iteration as a greatest fixed point}
Our viability kernel $\K$ is computed by iterating a monotone operator from the top safe set until convergence (Section~\ref{sec:engine}). In finite state spaces, this process stabilizes and yields the \emph{greatest} fixed point among all fixed points of the operator. In our finite kernel setting the controlled-invariance operator is explicitly contracting ($\mathcal{V}(K)\subseteq K$), matching the $\forall s,\ F(s)\subseteq s$ hypothesis in the Lean theorem. We provide a lightweight formal anchor of this fact in Lean 4 \citep{demoura2021lean4,mathlib2019}: see Appendix~\ref{app:lean_viability} and the corresponding file \texttt{lean/Agency/Viability.lean} in the repository.

\section{Discussion: an agent is a \emph{theory object}}
\label{sec:discussion}

We can now restate the main claim in the most compressed form, with the typing made explicit.

\paragraph{Thesis.}
In \SBT\ (\citep{six_birds_theory}), a \emph{theory} is a \emph{layer/closure}: an induced macro-physics, often written $T=(\Pi,L,\mathcal F,B)$. An agent is \emph{not} this layer. An agent is a \emph{theory object}: an object \emph{inside} a theory---a maintained package with a ledger-gated interface whose internal degrees of freedom become bona fide causes of external futures \emph{within} the induced macro-physics. Agency is the within-layer causal content of that object; agenthood is the cross-layer enablement that makes such an object exist and persist.

This framing matches the \SBT\ program: rather than positing agency as an extra ingredient, we ask which emergence primitives must be present for ``throwing a stone'' (stable counterfactual difference-making under constraints) to become a meaningful statement at an induced scale.

\subsection{Agenthood versus agency, revisited}
Section~\ref{sec:introduction} separated two often-confused notions.

\textbf{Agenthood} is an enablement claim: a theory/layer exists and persists long enough that an action variable can be identified at the boundary, a ledger can gate feasibility, and a safe set can be maintained robustly. In our finite setting, the viability kernel $\K$ operationalizes this persistence: if $\K=\emptyset$, the layer does not induce any maintained theory object on which policy choice can be meaningfully discussed as a stable capacity.

\textbf{Agency} is a causal claim inside the induced layer: interventions on the action variable change the distribution of future outside macrostates. In our setting, feasible empowerment is a clean proxy for this difference-making: it is the capacity of the induced channel from feasible action sequences to an outside output lens.

The exhibits intentionally split these axes. Repair can stabilize packaging (defect collapse) even when empowerment is limited; protocol holonomy can increase empowerment without changing the viability kernel size; and null regimes show that empowerment can be spuriously inflated if one misidentifies exogenous structure as choice.

\subsection{Causation versus enablement}
The \SBT\ distinction between causation and enablement is not philosophical decoration; it is the technical hinge of the paper.

Enablement is about \emph{making variables and objects exist} in a stable way: packaging produces an induced macrostate and boundary interface; accounting produces a ledger that can be written and spent; constraints produce a feasibility gate that turns ``commands'' into ``actions''; staging produces horizons on which endomaps can stabilize. These are cross-layer statements: they describe how a theory/layer becomes available and how it induces stable theory objects.

Causation is about \emph{difference-making once the variables exist}. Inside a fixed induced layer, we can meaningfully ask whether changing $a_t$ changes the distribution of $y_{t+H}$. Feasible empowerment is one proxy for that causal leverage. The schedule trap demonstrates why the separation matters: if we smuggle enablement mistakes into the causal layer (by treating an exogenous schedule as action), we fabricate agency.

\subsection{Relation to the Life paper}
The Life paper (\citep{six_birds_life}) treated life as a closure-and-maintenance phenomenon: what matters is not a biological checklist but the existence of a maintained package with accounting and persistence. The present paper extends that framing by adding a third ingredient that is not required for life as such: \emph{a nontrivial controllable interface whose counterfactual choices propagate to outside futures in a stable, compressible way}.

In short: life emphasizes P$_5$ (closure) and P$_6$ (accounting/maintenance) as the backbone of persistence; agency adds the requirement that the induced package supports a genuine action channel (difference-making), often amplified by P$_3$ (protocol holonomy) and strengthened by P$_1$ (operator rewriting). This aligns with everyday intuitions: many living systems persist without rich horizon-dependent control, and many controlled systems can act but do not persist autonomously. The six-birds dictionary disentangles these cases without importing goal talk.

\subsection{Limitations and failure modes}
\paragraph{Toy substrate and finite-state scope.}
Our environment is intentionally small and exact. This buys auditability and eliminates estimation ambiguity, but it is not a claim that agency in the wild reduces to small kernels. The correct interpretation is: the definitions are scale-agnostic, while the exhibits are minimal witnesses.

\paragraph{Lens dependence (what counts as outside).}
Empowerment depends on the chosen output lens $f$ and packaging depends on the macro lens $\pi$. This is not a bug: it reflects the core \SBT\ idea that macroscopic variables are induced. But it implies that agency statements are always relative to a description. We mitigated this by using stable, explicitly defined lenses (outside position $y$; macro lens $(y,r,\phi)$) and by including null regimes that catch obvious mis-modeling.

\paragraph{Empowerment is not a goal theory.}
Channel capacity is a proxy for difference-making, not a statement about preferences, utility, or optimal behavior. An agent can have high empowerment and still be ``aimless''; conversely, a system can be highly goal-directed in a narrow channel with low empowerment. This paper deliberately avoids importing goals as primitives.

\paragraph{Robust support semantics are conservative.}
We defined viability using successor support inclusion (all nonzero-probability outcomes must remain safe). This is appropriate for robust closure claims, but it can be stricter than expected-value safety. Different application domains may prefer risk-sensitive variants; the formal structure (greatest fixed point of a monotone operator) remains, but the safe predicate and post operator change.

\paragraph{Primitive coverage is uneven.}
P$_3$ (protocol) and P$_6$ (maintenance) are strongly exhibited; P$_1$ (rewrite) is demonstrated in a stylized way; P$_4$ (quantized identity/staging) is present as a discrete sector and horizon alignment but not deeply ablated as a separate commitment/identity phenomenon. A fuller treatment of P$_4$ would require richer commitments and multi-agent interactions.

\paragraph{Single-agent focus and absence of norms.}
We did not model social agency, bargaining, institutionally enforced constraints, or normativity. In later \SBT\ layers, P$_2$ constraints can encode laws and norms, and P$_4$ can encode identity/commitment tokens; those deserve a dedicated treatment.

\paragraph{Sampling and scale.}
Some reported values (e.g., median empowerment on $\K$) depend on deterministic sampling when $\K$ is large. This is acceptable for a toy witness but must be replaced by principled aggregation (or bounds) in large-scale settings.

\paragraph{Interface is assumed, not discovered.}
We treat the interface/action alphabet as part of the induced theory layer. Discovering boundaries and controllable interfaces from microdynamics is itself a packaging problem (a theory-construction problem) and is out of scope for this minimal witness.

\paragraph{Ledger is an abstract resource.}
The ledger variable $r$ is an accounting device used to model feasibility and maintenance costs. We do not claim thermodynamic optimality or a specific physical interpretation here; connecting $r$ to concrete energy/information inequalities is a separate layer construction.

\paragraph{What we did not claim.}
We did \emph{not} claim that agency requires consciousness, that empowerment is the only or final measure of agency, that our toy ring-world captures the richness of real organisms, or that an ``agent'' must have goals or utilities. We claimed something narrower and more structural: once an induced theory layer exists and induces a maintained theory object under budgets, agency is the existence of a genuine, model-respecting action channel that makes stable counterfactual differences at the relevant horizon.

\subsection{Outlook}
Three directions follow naturally. First, extend the substrate to multi-agent settings where P$_2$ constraints and P$_4$ tokens can encode commitments, ownership, and norms. Second, move from exact kernels to approximate learned models while preserving the audit posture (hashing, manifests, and invariants). Third, explore alternative causal proxies (risk-sensitive empowerment, reachability volumes, intervention-based causal effect sizes) and characterize when they agree or disagree.

\paragraph{Closing: the stone, the birds, and the theory.}
To throw a stone is to have a controllable interface whose choices propagate to the outside in a stable way under constraints. The six birds explain how such a statement can become true: packaging and viability make an induced layer persist; accounting and constraints define what actions exist; protocol holonomy and operator rewriting enrich what those actions can do. In that sense, an agent is a \emph{theory object}---not the layer itself, and not a story in its head, but a maintained object induced by a layer that turns certain internal degrees of freedom into causes in the induced macro-physics.

\appendix
\section{Lean anchor: viability iteration computes the greatest fixed point}
\label{app:lean_viability}

The viability kernel in Section~\ref{sec:engine} is computed as the limit of iterating a monotone, contracting operator on a finite lattice of sets. In a finite setting, iterating from the top element stabilizes at a fixed point, and this fixed point is greatest among all fixed points (a finite Tarski-style result) \citep{tarski1955}.

We formalize this statement in Lean 4 (mathlib) for operators on \texttt{Finset} with decidable equality and finiteness assumptions \citep{demoura2021lean4,mathlib2019}. The main exported theorem in \texttt{lean/Agency/Viability.lean} is:

\begin{verbatim}
theorem iterate_top_greatest_fixpoint
  (F : Finset α → Finset α) (hmono : Monotone F) (hsub : ∀ s, F s ⊆ s) :
  ∃ n : Nat,
    let K := Nat.iterate F n Finset.univ
    F K = K ∧ ∀ S : Finset α, F S = S → S ⊆ K
\end{verbatim}

In words: for any monotone operator $F$ that is pointwise contracting ($F(s)\subseteq s$), some iterate of $F$ applied to the top element (\texttt{univ}) yields a fixed point $K$, and every other fixed point $S$ is a subset of $K$.

\clearpage
\bibliographystyle{plainnat}
\bibliography{bib/refs}
\end{document}